\documentclass[lettersize,journal]{IEEEtran}
\usepackage{array}
\usepackage{colortbl}
\usepackage[caption=false,font=normalsize,labelfont=sf,textfont=sf]{subfig}
\usepackage{textcomp}
\usepackage{stfloats}
\usepackage{url}
\usepackage{verbatim}
\usepackage{graphicx}
\usepackage{cite}
\usepackage{placeins}
\newcommand*{\belowrulesepcolor}[1]{%
  \noalign{%
    \kern-\belowrulesep 
    \begingroup 
      \color{#1}%
      \hrule height\belowrulesep 
    \endgroup 
  }%
}

\newcommand*{\aboverulesepcolor}[1]{%
  \noalign{%
    \begingroup 
      \color{#1}%
      \hrule height\aboverulesep 
    \endgroup 
    \kern-\aboverulesep 
  }%
}

\usepackage{multirow}
\usepackage{bbding}
\usepackage{booktabs}
\usepackage{hyperref}
\usepackage{float}
\usepackage{mathrsfs}
\usepackage{orcidlink}
\usepackage{amssymb}
\usepackage{amsmath}
\usepackage{booktabs}

\definecolor{highlightblue}{RGB}{226,243,250}
\usepackage{graphicx}

\begin{document}

\title{VLA-ReID: Video-Level Association for Re-Identification in Multi-Object Tracking with Highly Similar Objects}
\author{
        Yanrong Qin, 
        Xiaoyan Cao\orcidlink{0000-0002-4980-2050},
        Yao Yao\orcidlink{0000-0001-9887-4301}
\thanks{
\textit{Corresponding author: Yao Yao.}}
\thanks{
Yanrong Qin is with Glasgow College, University of Electronic Science and Technology of China, Chengdu, China (e-mail: 2960033Q@student.gla.ac.uk).}
\thanks{
Xiaoyan Cao is with Key Laboratory for Urban Habitat Environmental Science and Technology, School of Environment and Energy, Peking University Shenzhen Graduate School, Shenzhen, China (e-mail: caoxiaoyan@stu.pku.edu.cn).}
\thanks{
Yao Yao is with School of Management Science and Real Estate, Chongqing University, Chongqing, China (e-mail: yao\_yao@cqu.edu.cn).
}
}

\markboth{Journal of \LaTeX\ Class Files,~Vol.~14, No.~8, August~2021}%
{Shell \MakeLowercase{\textit{et al.}}: A Sample Article Using IEEEtran.cls for IEEE Journals}
\maketitle

\begin{abstract}
Multi-object tracking (MOT) aims to continuously localize multiple objects in video streams while preserving their identities over time. However, long-term identity preservation remains difficult when objects are small, densely distributed, and highly similar in appearance, as in bee swarm scenes. Existing trackers rely on re-identification (re-ID) models to learn appearance representations for identity discrimination, but these models are typically trained through single-instance assignment (instance-level querying). At inference, however, data association in MOT requires global assignment between multiple trajectories and multiple detections (video-level querying). This mismatch between training and inference tends to cause identity switches among highly similar objects. Moreover, existing studies enhance appearance features to alleviate the difficulty of distinguishing highly similar objects, but typically require large amounts of additional annotated data. To address these issues, we propose Video-Level Association re-ID (VLA-ReID), which reformulates re-ID from instance-level training to video-level association modeling. Specifically, it treats aggregated historical trajectory features as queries and all detections in the current frame as candidates, thereby enabling direct optimization of the global association between them at each frame. Furthermore, Frame-Common Appearance Estimation (FCAE) estimates the common appearance direction from all detections in the current frame, while Common-Appearance Suppression (CAS) subtracts the component along this direction from both trajectory and detection features, thereby amplifying the discriminative differences among highly similar objects without additional annotations. Extensive experiments on BEE24 demonstrate that VLA-ReID outperforms state-of-the-art trackers in terms of HOTA (+1.1), MOTA (+0.3), AssR (+2.6), AssA (+0.7), and IDF1 (+0.8), while substantially reducing identity switches by 28\%. These results suggest that the video-level reformulation of re-ID offers a new perspective on appearance modeling in MOT. The source code and dataset are available at https://github.com/holmescao/VLA-ReID.
\end{abstract}


\section{Introduction}
\label{sec:introduction}
\IEEEPARstart{A}{s} a fundamental task in computer vision, multi-object tracking (MOT) has been widely applied in scenarios such as intelligent video analysis, traffic monitoring, and animal behavior research~\cite{markerlessbee,idtrackerai,anttracking,antdataset,swarmtracking,beemonitoring}. MOT aims to continuously localize multiple objects in a video stream while preserving their identities over time. In recent years, considerable progress has been achieved in various scenarios, including pedestrian and vehicle tracking. However, reliable long-term identity preservation remains highly challenging in scenes where objects are small, densely distributed, and highly similar in appearance~\cite{dancetrack,bee24}. The BEE24 bee-tracking dataset~\cite{bee24} is representative of such scenarios: it requires tracking dozens of bees over video sequences of up to 5,000 frames, lasting several minutes. In these sequences, the bees exhibit highly similar colors, shapes, scales, and stripe patterns, while their frequent interactions cause substantial mutual occlusion.

\begin{figure*}[tb]
\centering
\includegraphics[width=0.93\textwidth]{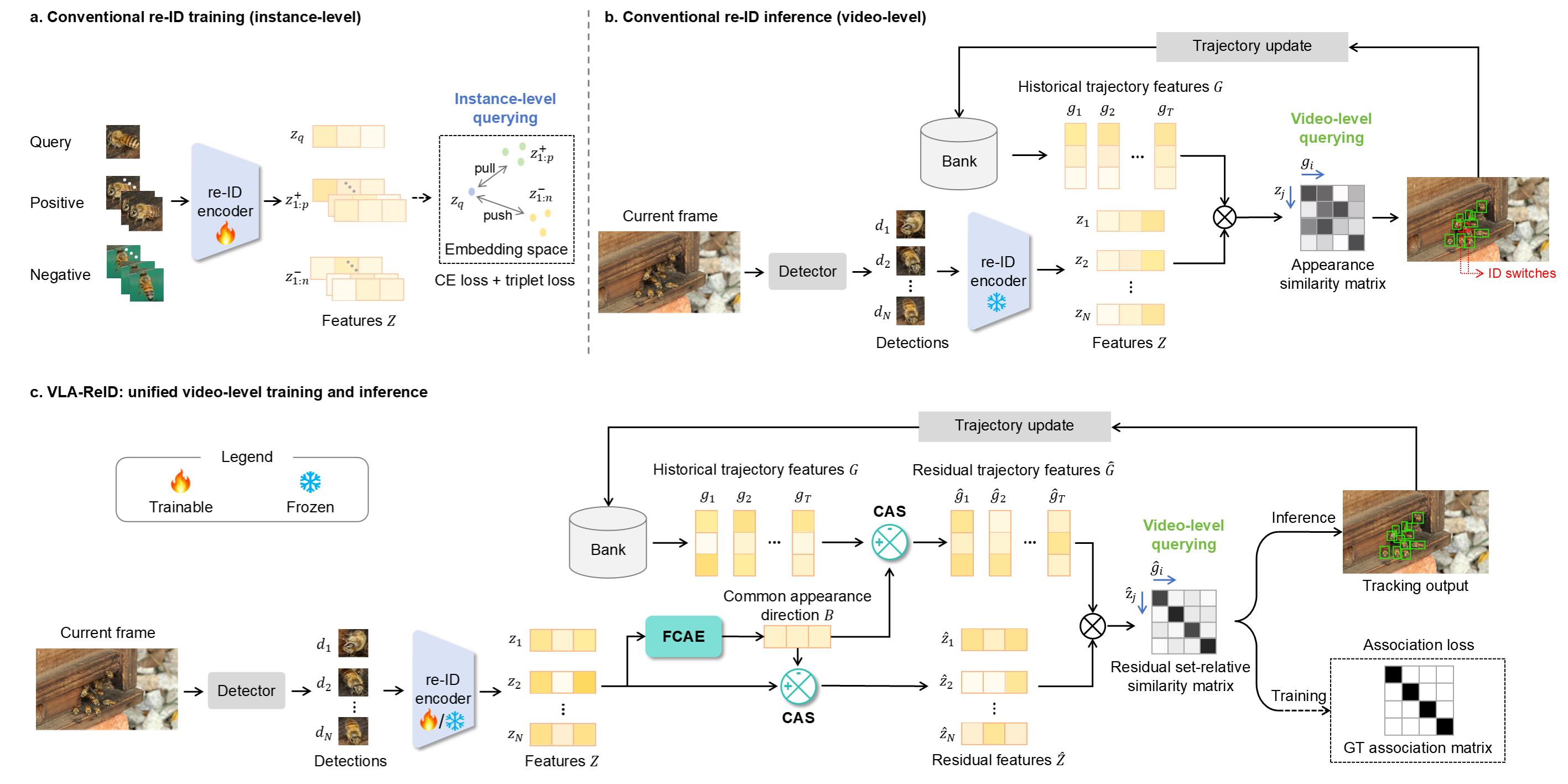}
\caption{Motivation and overview of VLA-ReID.
(a) Conventional re-ID training learns appearance embeddings through instance-level querying.
(b) During online MOT inference, historical trajectory features serve as video-level queries for joint global assignment with all current-frame detections. This mismatch between instance-level querying during training and video-level joint global assignment during inference can lead to ambiguous similarity scores and identity switches among highly similar objects.
(c) VLA-ReID aligns re-ID training with MOT inference through video-level candidate-set learning. FCAE estimates a common appearance direction from current-frame detections, while CAS suppresses the common appearance component along this direction in both trajectory and detection features. The resulting residual set-relative similarity matrix is optimized against the ground-truth (GT) association matrix during training and directly used for online data association during inference.
}
\label{fig:framework}
\end{figure*}

Tracking-by-detection (TBD) has become the dominant paradigm in MOT~\cite{sort,deepsort}. It consists of two main stages: (1) object detection, in which a detector generates candidate boxes for the current frame, and (2) data association, which assigns these detections to historical trajectories. Learning discriminative appearance representations is a key approach to data association~\cite{qdtrack}. Most appearance-based MOT methods rely on re-identification (re-ID) models, which were originally designed for image-based retrieval and matching. Such models are trained on batches of identity-annotated object patches randomly sampled from video sequences. For a given query patch, patches with the same identity are regarded as positive samples, whereas those with different identities serve as negative samples. The models then learn identity-discriminative representations at the instance level through identity classification or metric learning~\cite{adasp,dqal}, such as the triplet loss~\cite{hermans}, which pulls positive samples closer and pushes negative ones away (Fig.~\ref{fig:framework}a). 
In essence, conventional re-ID models optimize each query--candidate relationship independently, corresponding to single-instance assignment (instance-level querying). During MOT inference, however, data association requires joint global assignment between multiple historical trajectory features and all current-frame detections (video-level querying), where each historical trajectory feature aggregates observations across preceding frames~\cite{motip}. This training--inference mismatch can produce ambiguous association scores and increase the risk of identity switches among visually similar objects~\cite{deconfusetrack,mtcl,yao2021sample} (Fig.~\ref{fig:framework}b).

Existing re-ID studies~\cite{honeybeereid,bumblebeereid,openanimals,animalsimilarity} have mainly addressed the difficulty of distinguishing objects with highly similar appearances along two lines: (1) reducing the interference of identity-irrelevant nuisance factors~\cite{hanwooreid,detailenhancement} through preprocessing techniques such as pose or orientation normalization and background segmentation~\cite{pawvit}; and (2) extracting part-level or stripe-level local features~\cite{specieslocal} and fusing them with global features~\cite{multibranch} to highlight fine-grained individual differences~\cite{care}. However, these methods typically rely on additional supervision, often requiring extensive identity, pose, segmentation, or part-level annotations. Obtaining such annotations for large-scale video data is time-consuming and labor-intensive.

To address these issues, we propose Video-Level Association re-ID (VLA-ReID), in which the identity evidence for each trajectory is learned relative to the candidate set in the current frame. Specifically, VLA-ReID treats exponentially aggregated historical trajectory features as queries and all detections in the current frame as candidates, thereby enabling direct optimization of the complete trajectory-to-detection association matrix at each frame (Fig.~\ref{fig:framework}c). To further address the high appearance similarity among objects, we introduce two modules: Frame-Common Appearance Estimation (FCAE), which estimates a common appearance direction from all detections in the current frame, and Common-Appearance Suppression (CAS), which subtracts the common appearance component along this direction from both trajectory and detection features. This operation amplifies the discriminative differences among visually similar objects. Unlike the aforementioned methods, our approach estimates the common appearance direction online from current-frame detections and requires no additional annotations.

The main contributions of this work are summarized as follows:
\begin{itemize}
    \item We propose VLA-ReID, a novel video-level association modeling framework for MOT in scenarios involving highly similar objects. It reformulates instance-level re-ID learning as global association learning between the trajectory-feature set and the current-frame detection-feature set, thereby aligning re-ID training with the association process used during inference.
    \item We introduce a common-appearance estimation-and-suppression mechanism consisting of two modules. FCAE estimates the common appearance direction from current-frame detections, while CAS suppresses the common appearance component along this direction in both trajectory and detection features, thereby amplifying the discriminative differences among highly similar objects.
    \item We conduct a systematic evaluation on the BEE24 dataset. Compared with state-of-the-art trackers, VLA-ReID achieves the best performance on key metrics, including HOTA (+1.1), MOTA (+0.3), AssR (+2.6), AssA (+0.7), and IDF1 (+0.8), while reducing identity switches (IDs) by 28\%. Moreover, ablation studies and sensitivity analyses validate the effectiveness of the proposed components. These results demonstrate that our method effectively improves the quality of identity association in scenarios with highly similar objects.
\end{itemize}

\section{Method}
\label{sec:method}
\subsection{Overview}
\label{sec:method_overview}
Conventional re-ID training optimizes identity discrimination through instance-level querying, whereas online MOT inference requires joint association between trajectories and all detections in the current frame. VLA-ReID addresses this mismatch by learning identity evidence relative to the complete trajectory-to-detection candidate set.
As illustrated in Fig.~\ref{fig:framework}c, FCAE estimates a common appearance direction from current-frame detections, and CAS suppresses the gated component along this direction in both detection and historical trajectory features. The resulting residual features form a residual set-relative similarity matrix that emphasizes discriminative differences within the current candidate set.
During inference, this matrix is converted into a set-relative appearance cost and integrated into the base tracker without altering its other components. During training, the appearance encoder, FCAE, and CAS are jointly optimized with video-level episodes that reproduce the candidate-set competition encountered in association.

\subsection{Preliminaries and Notation}
\label{sec:preliminaries}

We build on TrackTrack~\cite{tracktrack}, an online tracking-by-detection method, while leaving its detector, motion model, matching procedure, and trajectory management unchanged. At frame \(t\), the detector produces \(N\) detections. Each detection is encoded by the AGW appearance encoder implemented in FastReID~\cite{fastreid,agw}, yielding an \(\ell_2\)-normalized \(D\)-dimensional feature, where \(D=2048\) under the standard configuration. The detection features are collected as
\begin{equation}
Z=[z_1;\ldots;z_N]\in\mathbb{R}^{N\times D}.
\label{eq:detection-features}
\end{equation}

Each active trajectory \(i\) maintains a historical trajectory feature \(g_i\). 
At inference, if trajectory \(i\) is successfully matched to detection \(j\), its feature is updated using a confidence-weighted exponential moving average, which provides a lightweight form of temporal state propagation~\cite{cao2026large,cao2025u}:
\begin{equation}
\begin{aligned}
g_i
&\leftarrow
\operatorname{normalize}\!\left(
\beta g_i+(1-\beta)z_j
\right),\\
\beta
&=
\mu+(1-\mu)(1-s_j),
\end{aligned}
\label{eq:trajectory-feature-update}
\end{equation}
where \(z_j\) and \(s_j\) denote the detection feature and confidence score of the matched detection \(j\), respectively. The base decay factor is fixed as \(\mu=0.95\). The \(T\) historical trajectory features are collected as
\begin{equation}
G=[g_1;\ldots;g_T]\in\mathbb{R}^{T\times D}.
\label{eq:trajectory-features}
\end{equation}

TrackTrack performs two association calls per frame. The first stage matches active and lost trajectories to detections, whereas the second stage matches newly initialized trajectories to the remaining high-confidence detections. Each call performs association using the following association cost matrix:
\begin{equation}
C
=
\lambda_{\mathrm{iou}}C^{\mathrm{iou}}
+
\lambda_{\mathrm{app}}C^{\mathrm{app}}
+
\lambda_{\mathrm{conf}}C^{\mathrm{conf}}
+
\lambda_{\mathrm{ang}}C^{\mathrm{ang}},
\label{eq:base-cost}
\end{equation}
where
\(C^{\mathrm{iou}}\),
\(C^{\mathrm{app}}\),
\(C^{\mathrm{conf}}\),
and
\(C^{\mathrm{ang}}\)
denote the IoU, appearance, confidence, and angular cost terms, respectively. The corresponding weights are inherited from TrackTrack and fixed as
\[
\lambda_{\mathrm{iou}}=0.50,\quad
\lambda_{\mathrm{app}}=0.50,\quad
\lambda_{\mathrm{conf}}=0.10,\quad
\lambda_{\mathrm{ang}}=0.05.
\]
The pairwise appearance cost is defined as
\begin{equation}
C^{\mathrm{app}}_{ij}
=
\operatorname{clip}\!\left(
1-g_i^\top z_j,\,
0,\,
1
\right),
\label{eq:app-distance}
\end{equation}
whereas the other cost terms are computed identically to the original TrackTrack implementation~\cite{tracktrack}. Low-confidence and NMS-recovered detections receive additive penalties, and trajectory--detection pairs with an IoU similarity of \(0.1\) or lower are excluded from association. Consequently, appearance matching is performed only among spatially plausible trajectory--detection pairs. The proposed method reformulates only the pairwise appearance cost, leaving all remaining components of the association cost unchanged.

Pairwise appearance comparison can be unreliable in dense same-species scenes. Co-occurring bees often share similar body patterns, scales, illumination conditions, and background remnants, whereas identity-specific differences between the most confusable same-frame candidates may be subtle. The discriminative value of an appearance cue therefore depends on the candidate set present in the current frame. This set-dependent ambiguity motivates our common-appearance estimation-and-suppression mechanism. FCAE estimates the common appearance direction from current-frame detections, while CAS suppresses the common appearance component along this direction in both detection and trajectory features before association.

\subsection{Frame-Common Appearance Estimation}
\label{sec:fcae}

\begin{figure*}[t]
\centering
\includegraphics[width=0.98\textwidth]{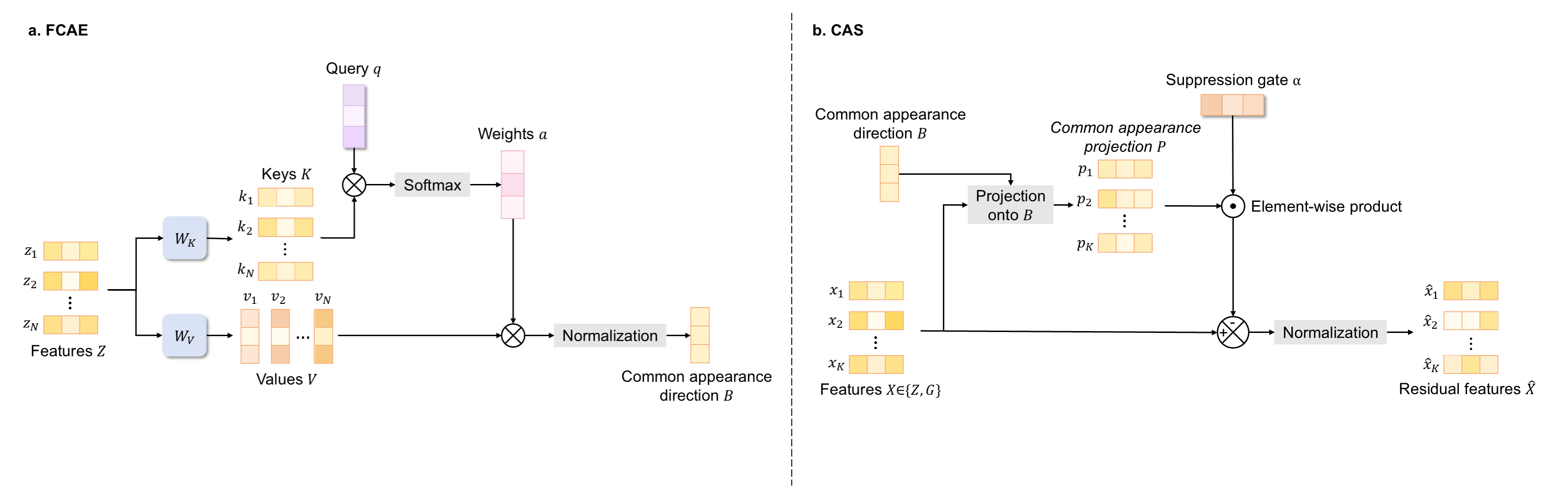}
\caption{Detailed workflow of FCAE and CAS. FCAE estimates the common appearance direction \(B\) from the current detection feature set \(Z\). Given either detection features or historical trajectory features, \(X\in\{Z,G\}\), CAS applies the same direction \(B\) and learnable suppression gate \(\alpha\) to obtain the residual features \(\hat{X}\).}
\label{fig:fcae-cas}
\end{figure*}

FCAE estimates a common appearance direction from the current detection set. Because the common appearance is a property of the candidate set rather than of individual detections, FCAE estimates this direction using learned attention pooling. Given \(Z\in\mathbb{R}^{N\times D}\), a learnable query \(q\in\mathbb{R}^{1\times D}\) attends to the detection features (see Fig.~\ref{fig:fcae-cas}a):
\begin{align}
K &= ZW_K^\top,\quad
V = ZW_V^\top,
\nonumber\\
a
&=
\operatorname{softmax}\!\left(
\frac{qK^\top}{\sqrt{D}\,\tau_f}
\right)
\in\mathbb{R}^{1\times N},
\label{eq:fcae-attention}
\\
B
&=
\frac{aV}{\lVert aV\rVert_2}
\in\mathbb{R}^{1\times D},
\label{eq:fcae-direction}
\end{align}
where \(W_K,W_V\in\mathbb{R}^{D\times D}\) are learned linear projections, \(a\) denotes the attention weights over the \(N\) detections, and \(\tau_f=1\) is the attention temperature. The unit-norm vector \(B\) is the estimated common appearance direction.

Attention pooling is permutation-invariant with respect to the candidate set and adaptively estimates the common appearance direction according to the current detections. For ablation, we replace attention pooling with the normalized set mean,
\begin{equation}
B_{\mathrm{mean}}
=
\operatorname{normalize}\!\left(
\frac{1}{N}
\sum_{j=1}^{N}
z_j
\right).
\label{eq:b_mean}
\end{equation}

All experiments use a single common appearance direction. Restricting suppression to one direction limits the removal of identity-bearing information while remaining sufficient to capture the dominant appearance commonality shared by the current candidate set.

\subsection{Common-Appearance Suppression}
\label{sec:cas}
CAS suppresses the common appearance component along the estimated common appearance direction and then re-normalizes the resulting feature. As shown in Fig.~\ref{fig:fcae-cas}b, for any \(\ell_2\)-normalized feature \(x\), which may be a detection feature \(z_j\) or a historical trajectory feature \(g_i\), CAS is defined as
\begin{equation}
\begin{aligned}
\hat{x}&=\operatorname{normalize}\!\left(
x-\alpha\odot(xB^\top)B
\right),\\
\alpha&=\sigma(\theta),\quad \theta\in\mathbb{R}^{1\times D},
\end{aligned}
\label{eq:cas}
\end{equation}
where \((xB^\top)B\) is the projection of \(x\) onto the common appearance direction \(B\), \(\odot\) denotes element-wise multiplication, \(\sigma(\cdot)\) is the sigmoid activation, and \(\theta\) is a learnable parameter vector shared across all samples. The resulting \(\alpha\in(0,1)^{1\times D}\) acts as a learnable per-dimension suppression gate~\cite{yao2023razor}.
When \(\alpha\rightarrow\mathbf{0}\), CAS degenerates to the identity mapping; when \(\alpha=\mathbf{1}\), the common appearance component is removed completely. We initialize \(\theta\) to a positive constant so that \(\alpha=\sigma(\theta)\) is initially close to \(\mathbf{1}\), and allow optimization to reduce the suppression strength in dimensions where it is detrimental.

The same common appearance direction \(B\) and suppression gate \(\alpha\) are applied to both detection and historical trajectory features:
\begin{equation}
\hat{Z}=\operatorname{CAS}(Z,B,\alpha),
\quad
\hat{G}=\operatorname{CAS}(G,B,\alpha).
\label{eq:cas-both-sides}
\end{equation}
This shared transformation projects queries and candidates into the same residual appearance space. The common appearance direction is estimated exclusively from the current detections because historical trajectory features aggregate observations across previous frames and would otherwise blur the shared appearance of the current candidate set.
The resulting residual set-relative similarity matrix is
\begin{equation}
S=\hat{G}\hat{Z}^{\top}
\in\mathbb{R}^{T\times N}.
\label{eq:residual-similarity}
\end{equation}
Because \(\hat{G}\) and \(\hat{Z}\) are \(\ell_2\)-normalized, \(S\) is the cosine similarity matrix computed from the residual features. By suppressing the common appearance component, the resulting residual set-relative similarity matrix places greater emphasis on feature components that distinguish candidates within the current frame. The objective is therefore set-relative re-ranking rather than enlarging global appearance margins.

\subsection{Set-Relative Re-scoring for Association}
\label{sec:rescoring}

For an association call with \(T\) trajectories and \(N\) candidate detections, the residual set-relative similarity matrix \(S\) is converted into a set-relative appearance cost by min--max normalization within the current score block:
\begin{align}
C^{\mathrm{set}}_{ij}
&=
1-
\frac{S_{ij}-S_{\min}}
{S_{\max}-S_{\min}+\varepsilon},
\label{eq:set-relative-distance}
\\
S_{\min}
&=
\min_{i,j} S_{ij},
\quad
S_{\max}
=
\max_{i,j} S_{ij},
\nonumber
\end{align}
where \(\varepsilon=10^{-6}\) is a small constant for numerical stability. The resulting set-relative appearance cost reflects how each trajectory--detection pair compares with the competing pairs in the current association call. Its numerical scale is therefore specific to the current candidate set. Spatial gating, the overall cost threshold, and the remaining tracker terms continue to constrain the assignment.
The final association cost is obtained by replacing the pairwise appearance cost in Eq.~\eqref{eq:base-cost} with the proposed set-relative appearance cost:
\begin{equation}
C
=
\lambda_{\mathrm{iou}}C^{\mathrm{iou}}
+
\lambda_{\mathrm{app}}C^{\mathrm{set}}
+
\lambda_{\mathrm{conf}}C^{\mathrm{conf}}
+
\lambda_{\mathrm{ang}}C^{\mathrm{ang}}.
\label{eq:final-cost}
\end{equation}

For sensitivity analysis, we introduce a blending coefficient \(c\) to interpolate between the original pairwise appearance cost and the proposed set-relative appearance cost:
\begin{equation}
\begin{aligned}
C(c)
={}&
\lambda_{\mathrm{iou}}C^{\mathrm{iou}}
+
\left(\lambda_{\mathrm{app}}-c\right)C^{\mathrm{app}}
+
cC^{\mathrm{set}}
\\
&+
\lambda_{\mathrm{conf}}C^{\mathrm{conf}}
+
\lambda_{\mathrm{ang}}C^{\mathrm{ang}},
\quad
c\in[0,\lambda_{\mathrm{app}}].
\end{aligned}
\label{eq:blended-cost}
\end{equation}
When \(c=0\), the association cost uses only the original pairwise appearance term; when \(c=\lambda_{\mathrm{app}}\), the pairwise appearance term is completely replaced by the set-relative appearance cost. The detector, motion model, penalties, spatial gating, matching procedure, and trajectory management remain unchanged. Set-relative re-scoring is applied in both association stages.

\begin{table*}[t]
\centering
\caption{Ablation on the BEE24 test set. All variants share the same detections and TrackTrack pipeline and differ only in the appearance branch. \emph{Baseline}: original cosine appearance distance. \emph{+Episode}: trajectory-to-frame candidate-set association training. \emph{+Mean+CAS}: adds a parameter-free set-mean commonality-suppression control. \emph{+FCAE+CAS}: jointly learned commonality suppression (final model). Best and second-best results in each column are shown in \textbf{bold} and \underline{underlined}, respectively.}
\label{tab:ablation}
\footnotesize
\setlength{\tabcolsep}{5pt}
\begin{tabular}{l|ccccccccc}
\toprule
Configuration
& HOTA \(\uparrow\)
& MOTA \(\uparrow\)
& AssR \(\uparrow\)
& AssA \(\uparrow\)
& IDF1 \(\uparrow\)
& FP \(\downarrow\)
& FN \(\downarrow\)
& IDs \(\downarrow\)
& Frag \(\downarrow\) \\
\midrule
Baseline (TrackTrack)
& 48.1 & 65.7 & 54.9 & 45.9 & 64.1
& \textbf{6026} & 51093 & 606 & 3925 \\
Baseline + Episode
& 48.4 & \textbf{67.5} & \textbf{61.9} & 44.4 & 63.1
& 14856 & \textbf{39283} & \underline{466} & \textbf{3345} \\
Baseline + Episode + Mean + CAS
& \underline{48.6} & 66.6 & 57.5 & \underline{46.0} & \underline{64.3}
& \underline{8409} & 47231 & 542 & 3738 \\
\midrule
\rowcolor{highlightblue}
Baseline + Episode + FCAE + CAS
& \textbf{49.2} & \underline{67.0} & \underline{61.7} & \textbf{46.6} & \textbf{64.9}
& 11956 & \underline{43096} & \textbf{437} & \underline{3462} \\
\bottomrule
\end{tabular}
\end{table*}

\begin{table*}[t]
\centering
\caption{
Tracking performance on the BEE24 test set. Methods in the shaded region use the same shared detections and are therefore directly comparable, whereas methods in the unshaded top block use their own detections. \(\uparrow\) and \(\downarrow\) indicate that higher and lower values are better, respectively. Within the shared-detection region, bold and underlined values denote the best and second-best results in each column, respectively.
}
\label{tab:sota}
\footnotesize
\setlength{\tabcolsep}{5pt}
\begin{tabular}{l|ccccccccc}
\toprule
Method
& HOTA \(\uparrow\)
& MOTA \(\uparrow\)
& AssR \(\uparrow\)
& AssA \(\uparrow\)
& IDF1 \(\uparrow\)
& FP \(\downarrow\)
& FN \(\downarrow\)
& IDs \(\downarrow\)
& Frag \(\downarrow\) \\
\midrule
CTracker~\cite{ctracker}    & 33.4 & 42.8 & 34.0 & 24.5 & 39.4 & 65532 & 23983 & 2987 & 2770 \\
FairMOT~\cite{fairmot}     & 42.3 & 40.9 & 57.6 & 42.5 & 54.3 & 75799 & 18501 & 3968 & 3790 \\
TraDeS~\cite{trades}      & 30.9 & 42.2 & 26.6 & 20.2 & 34.8 & 56966 & 26286 & 5660 & 4716 \\
TrackFormer~\cite{trackformer} & 44.3 & 41.5 & 55.5 & 42.3 & 53.9 & 86777 & 9989 & 3405 & 2271 \\
\midrule
\belowrulesepcolor{highlightblue}
\rowcolor{highlightblue} UniTrack~\cite{unitrack}          & 41.6 & 54.6 & 56.2 & 34.8 & 53.0 & 51369 & \textbf{21953} & 1972 & \textbf{2395} \\
\rowcolor{highlightblue} ByteTrack~\cite{bytetrack}         & 43.2 & 59.2 & 52.7 & 38.3 & 56.8 & 23343 & 44130 & 1303 & 6663 \\
\rowcolor{highlightblue} OC-SORT~\cite{ocsort}           & 42.7 & 61.6 & 50.8 & 36.8 & 55.3 & 20493 & 43172 & 1435 & 4996 \\
\rowcolor{highlightblue} TOPICTrack~\cite{bee24}        & 46.6 & \underline{66.7} & \underline{59.1} & 40.3 & 59.7 & 29171 & \underline{25691} & 1401 & \underline{2490} \\
\rowcolor{highlightblue} TrackTrack~\cite{tracktrack}        & \underline{48.1} & 65.7 & 54.9 & \underline{45.9} & \underline{64.1} & \textbf{6026} & 51093 & \underline{606} & 3925 \\
\midrule
\rowcolor{highlightblue} VLA-ReID (Ours)  & \textbf{49.2} & \textbf{67.0} & \textbf{61.7} & \textbf{46.6} & \textbf{64.9} & \underline{11956} & 43096 & \textbf{437} & 3462 \\
\aboverulesepcolor{highlightblue}
\bottomrule
\end{tabular}
\end{table*}

\subsection{Video-Level Episode Training}
\label{sec:episode_training}

To train VLA-ReID under the same candidate-set structure encountered during online inference, we construct video-level episodes centered on target frames. Each episode jointly presents historical trajectory features as queries and all detections in the target frame as candidates, allowing the appearance encoder, FCAE, and CAS to be optimized over the complete trajectory-to-detection association matrix. Here, ``video-level'' refers to the supervision structure: each query summarizes observations accumulated along a trajectory, although association remains online and is performed at the frame level. The model introduces no recurrent or temporal-attention module.

\noindent\textbf{Episode construction.}
An episode is anchored at a target frame \(t\) and contains the \(N\) detections in that frame together with \(T\) eligible trajectory histories whose identities are present at \(t\). Trajectories without a matched detection in the target frame, as well as decisions involving newly appearing objects, are excluded from the training objective; at inference, these cases remain governed by the thresholds and penalties of the base tracker. The detections are encoded into \(Z\), while each trajectory history is aggregated into a historical trajectory feature using an exponential moving average with a fixed decay factor of \(0.95\). The forward pass follows the inference path: the common appearance direction is estimated as \(B=\operatorname{FCAE}(Z)\), the residual features \(\hat{Z}\) and \(\hat{G}\) are computed using Eq.~\eqref{eq:cas-both-sides}, and the residual set-relative similarity matrix \(S\) is obtained from Eq.~\eqref{eq:residual-similarity}.

\noindent\textbf{Candidate-set competition.}
For each historical trajectory feature \(g_i\), let \(y_i\in\{1,\ldots,N\}\) denote the index of its corresponding detection in the target frame. The primary supervision is a row-wise candidate-set cross-entropy (CE) loss that requires each trajectory query to identify its matched detection among all current-frame candidates:
\begin{equation}
\mathcal{L}_{\mathrm{row}}
=
\frac{1}{T}
\sum_{i=1}^{T}
\operatorname{CE}\!\left(
\frac{S_{i,:}}{\tau_r},
y_i
\right),
\label{eq:row-loss}
\end{equation}
where \(\tau_r=0.1\) is the softmax temperature.

To enforce reciprocal trajectory--detection consistency, let \(\mathcal{J}\subseteq\{1,\ldots,N\}\) denote the set of detections with valid trajectory correspondences, and let \(r_j\in\{1,\ldots,T\}\) denote the index of the historical trajectory corresponding to detection \(j\). We then define the symmetric column-wise loss as
\begin{equation}
\mathcal{L}_{\mathrm{col}}
=
\frac{1}{|\mathcal{J}|}
\sum_{j\in\mathcal{J}}
\operatorname{CE}\!\left(
\frac{S_{:,j}}{\tau_r},
r_j
\right).
\label{eq:col-loss}
\end{equation}
Thus, \(\mathcal{L}_{\mathrm{row}}\) requires each trajectory query to select its matched detection, whereas \(\mathcal{L}_{\mathrm{col}}\) requires each valid detection to select its corresponding trajectory.

\noindent\textbf{Anti-collapse regularization.}
Optimizing candidate ranking alone may lead the estimated common appearance direction or the suppression gate to ineffective solutions. We therefore regularize both the separation between residual detection features and the alignment of \(B\) with the observed detection distribution:
\begin{equation}
\mathcal{L}_{\mathrm{com}}
=
w_{\mathrm{sep}}\,
\overline{\cos}
\bigl(\hat{z}_j,\hat{z}_{j'}\bigr)_{j\neq j'}
+
w_{\mathrm{anc}}
\left(
1-
\frac{1}{N}
\sum_{j=1}^{N}
\left|z_jB^\top\right|
\right). \nonumber
\label{eq:commonality-loss} 
\end{equation}
The first term is the mean off-diagonal cosine similarity between residual detection features and encourages greater separation among same-frame candidates. The second term anchors the common appearance direction \(B\) to the observed detection distribution, preventing FCAE from drifting toward a direction that is weakly represented in the detection features. We set \(w_{\mathrm{sep}}=1\) and \(w_{\mathrm{anc}}=0.1\).

The complete training objective is
\begin{equation}
\mathcal{L}
=
\mathcal{L}_{\mathrm{row}}
+
\lambda_{\mathrm{col}}\mathcal{L}_{\mathrm{col}}
+
\lambda_{\mathrm{com}}\mathcal{L}_{\mathrm{com}},
\label{eq:total-loss}
\end{equation}
where \(\lambda_{\mathrm{col}}=0.1\) and
\(\lambda_{\mathrm{com}}=0.5\) balance the column-wise candidate-set loss and the anti-collapse regularization, respectively.
To reduce memory consumption, encoder gradients are retained on the trajectory-query branch, whereas detection features are computed without gradients. During training, cross-entropy is applied directly to the unnormalized similarities in \(S\); min--max normalization is used only when the residual set-relative similarity matrix is converted into a set-relative appearance cost for online association.

\begin{figure}[tb]
\centering
\includegraphics[width=0.92\linewidth]{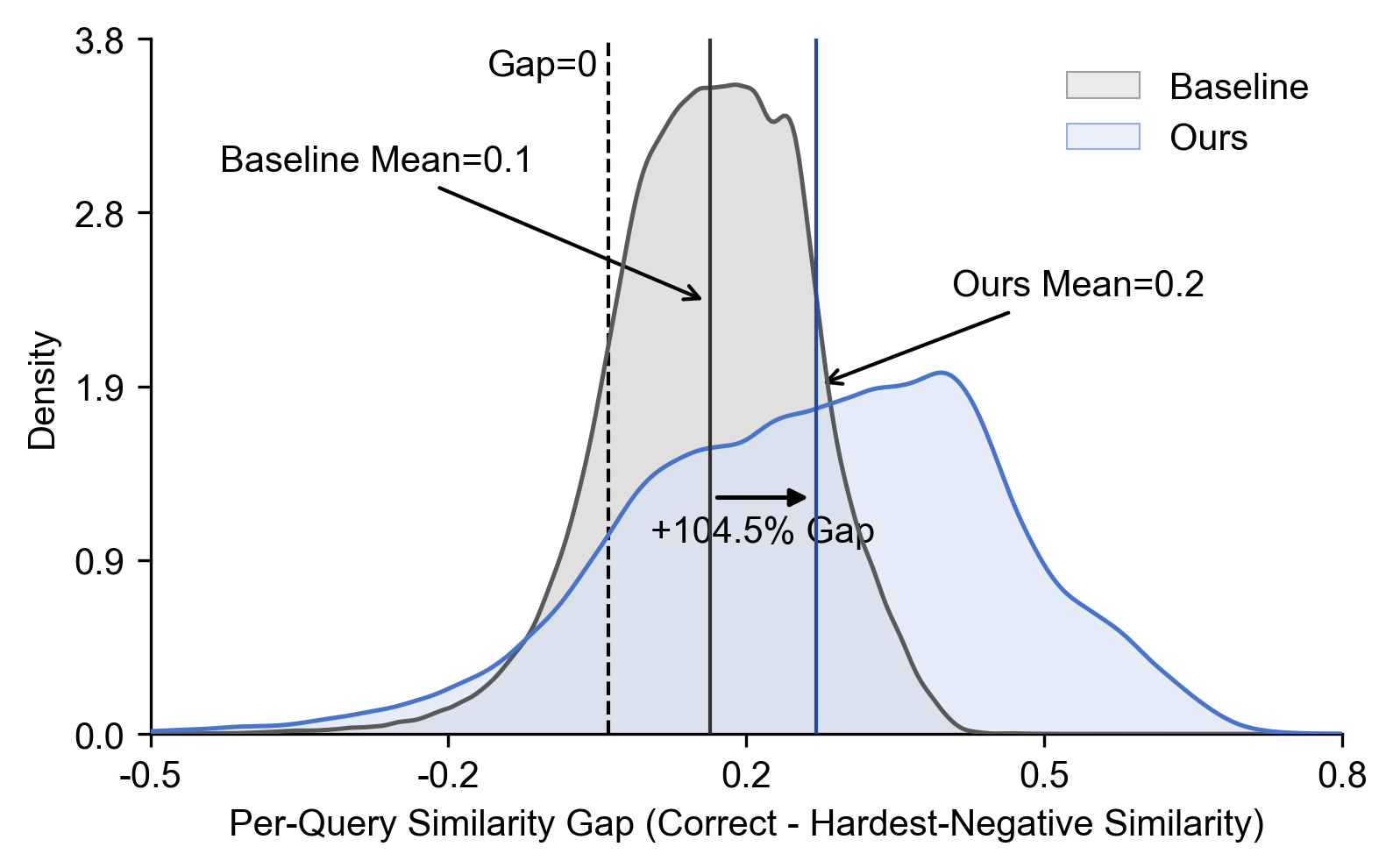}
\caption{Distribution of the per-query similarity gap between the true candidate and the highest-scoring incorrect candidate in the same frame. Larger values indicate better within-frame separation.}
\label{fig:similarity-gap}
\vspace{-1em}
\end{figure}
\begin{figure}[tb]
\centering
\includegraphics[width=0.92\linewidth]{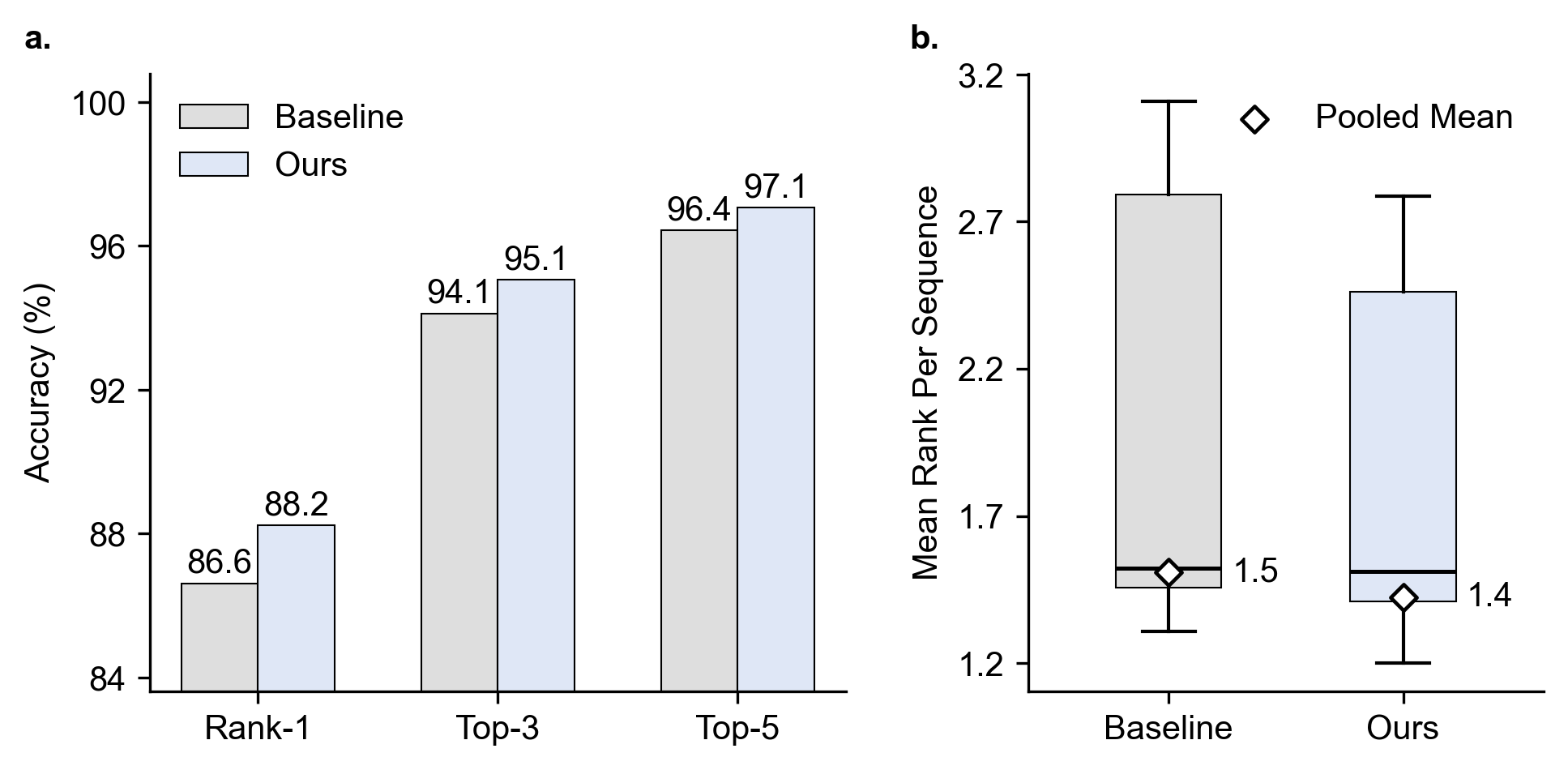}
\caption{Candidate-ranking performance. Rank-1, Top-3, and Top-5 report the proportion of queries for which the true candidate appears within the corresponding rank cutoff; lower mean rank is better.}
\label{fig:candidate-ranking}
\vspace{-1em}
\end{figure}
\begin{figure}[t!]
\centering
\includegraphics[width=0.92\linewidth]{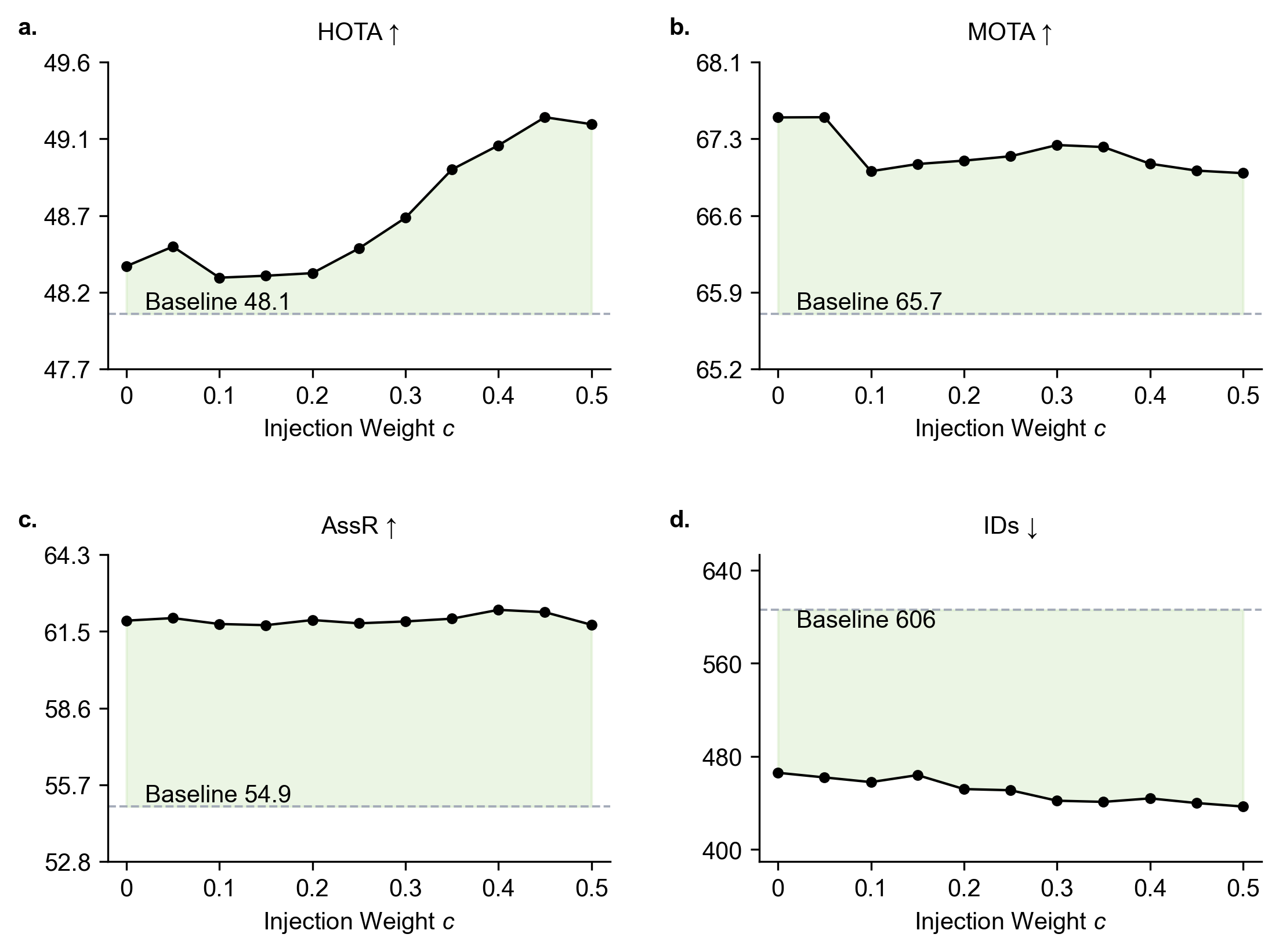}
\caption{Sensitivity to the injection weight \(c\). All non-appearance association terms are fixed; \(c\) controls the mixture of original pairwise appearance cost and the set-relative appearance cost.}
\label{fig:weight-sensitivity}
\vspace{-1em}
\end{figure}

\section{Experiments}
\label{sec:experiments}
\subsection{Implementation Details}
\label{sec:implementation}
We use YOLOX~\cite{yolox} as the detector and follow the shared-detection protocol of BEE24~\cite{bee24}, under which all compared trackers receive identical detections. The appearance encoder is the AGW model implemented in FastReID~\cite{agw,fastreid} and equipped with GeM pooling~\cite{gem}. We train the encoder from random initialization for \(120\) epochs using cross-entropy and triplet losses, with label smoothing of \(0.1\) and a triplet margin of \(0\). This setting avoids pretraining on pedestrian re-ID datasets.

We then construct video-level episodes from the re-ID training sequences. Each episode pairs historical trajectory features aggregated across preceding frames with the complete detection set of a target frame. The episode objective is used to fine-tune the appearance encoder and jointly train FCAE and CAS. Historical trajectory features are aggregated using an exponential moving average with a fixed decay factor of \(0.95\). We use AdamW~\cite{adamw} with learning rates of \(1\times10^{-4}\) for the appearance encoder and \(1\times10^{-3}\) for FCAE and CAS. The candidate-set temperature is set to \(\tau_r=0.1\), and episode training is conducted for \(3\) epochs on \(8\) NVIDIA RTX 3090 GPUs.

We integrate VLA-ReID into TrackTrack~\cite{tracktrack} by replacing only its pairwise appearance cost with the proposed set-relative appearance cost. All other components remain unchanged, including the NSA Kalman motion model~\cite{giaotracker}, Track-Perspective-Based Association, Track-Aware Initialization, the IoU, confidence, and angular cost terms, the matching procedure, and trajectory management.

\subsection{Evaluation Metrics}
\label{sec:metrics}

We evaluated tracking performance using HOTA, MOTA, AssR, AssA, IDF1, false positives (FP), false negatives (FN), identity switches (IDs), and fragmentations (Frag)~\cite{hota,clear-mot,idf1}. HOTA was the primary metric because it balances detection and association accuracy. MOTA summarizes FP, FN, and IDs. Its value is often dominated by detection errors.

Because this study focuses on preserving identity among highly similar targets, we emphasized IDF1 and the association metrics. IDF1 summarizes identity matching over a sequence. AssA measures the accuracy of associations for matched detections, whereas AssR measures the recovery of ground-truth association links. IDs count identity switches, and Frag counts interruptions of ground-truth trajectories. FP and FN characterize detection precision and recall within the tracking pipeline.

\begin{figure*}[!t]
\centering
\includegraphics[width=0.9\textwidth]{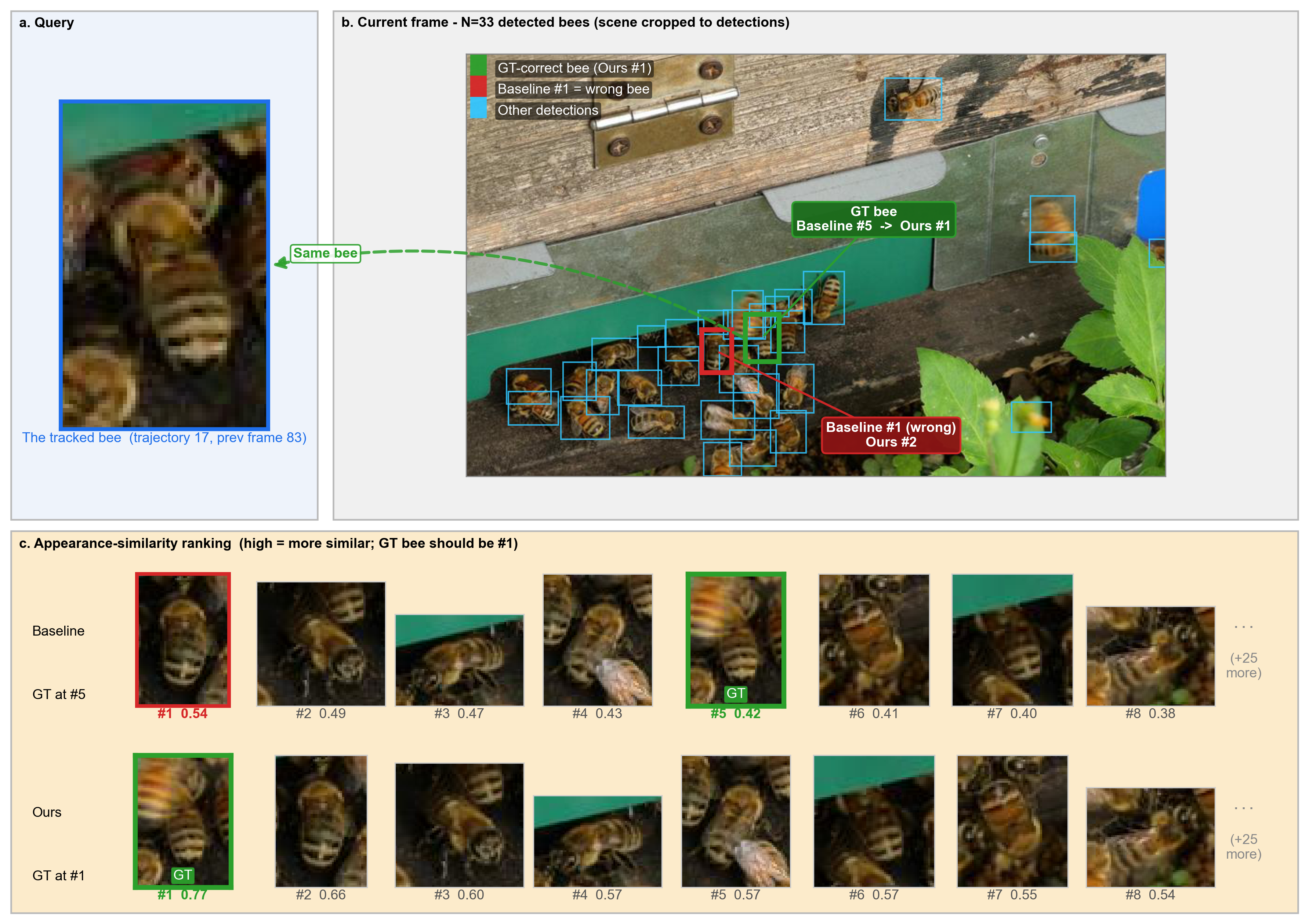}
\caption{Parameter-free commonality suppression moves the true candidate from rank 5 to rank 1 among 33 detections in a dense frame from BEE24-34.}
\label{fig:case-bee24-34}
\end{figure*}
\begin{figure*}[!th]
\centering
\includegraphics[width=0.93\textwidth]{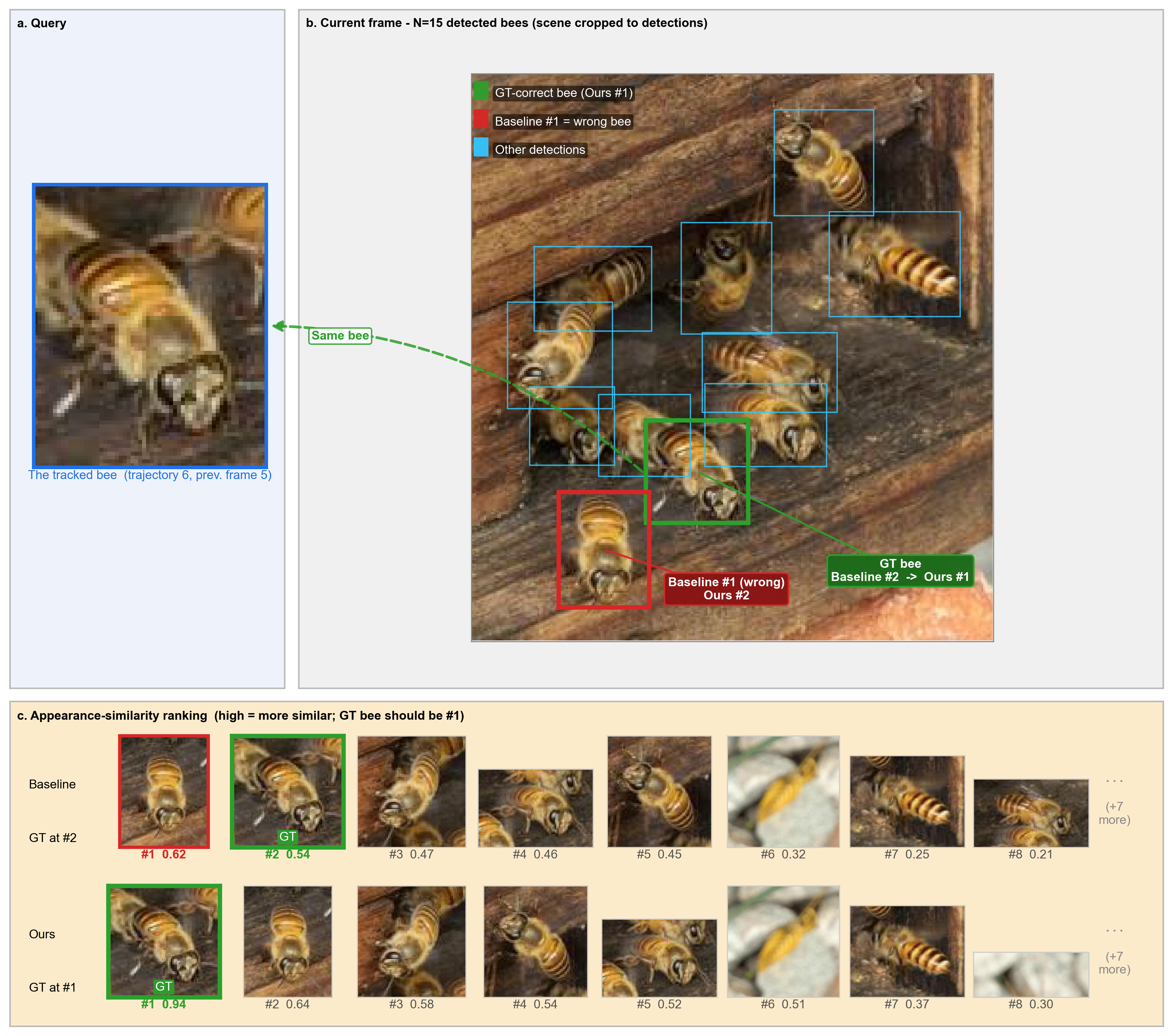}
\caption{Parameter-free commonality suppression moves the true candidate from rank 2 to rank 1 among 15 detections in BEE24-36.}
\label{fig:case-bee24-36}
\end{figure*}
\begin{figure*}[!th]
\centering
\includegraphics[width=0.92\textwidth]{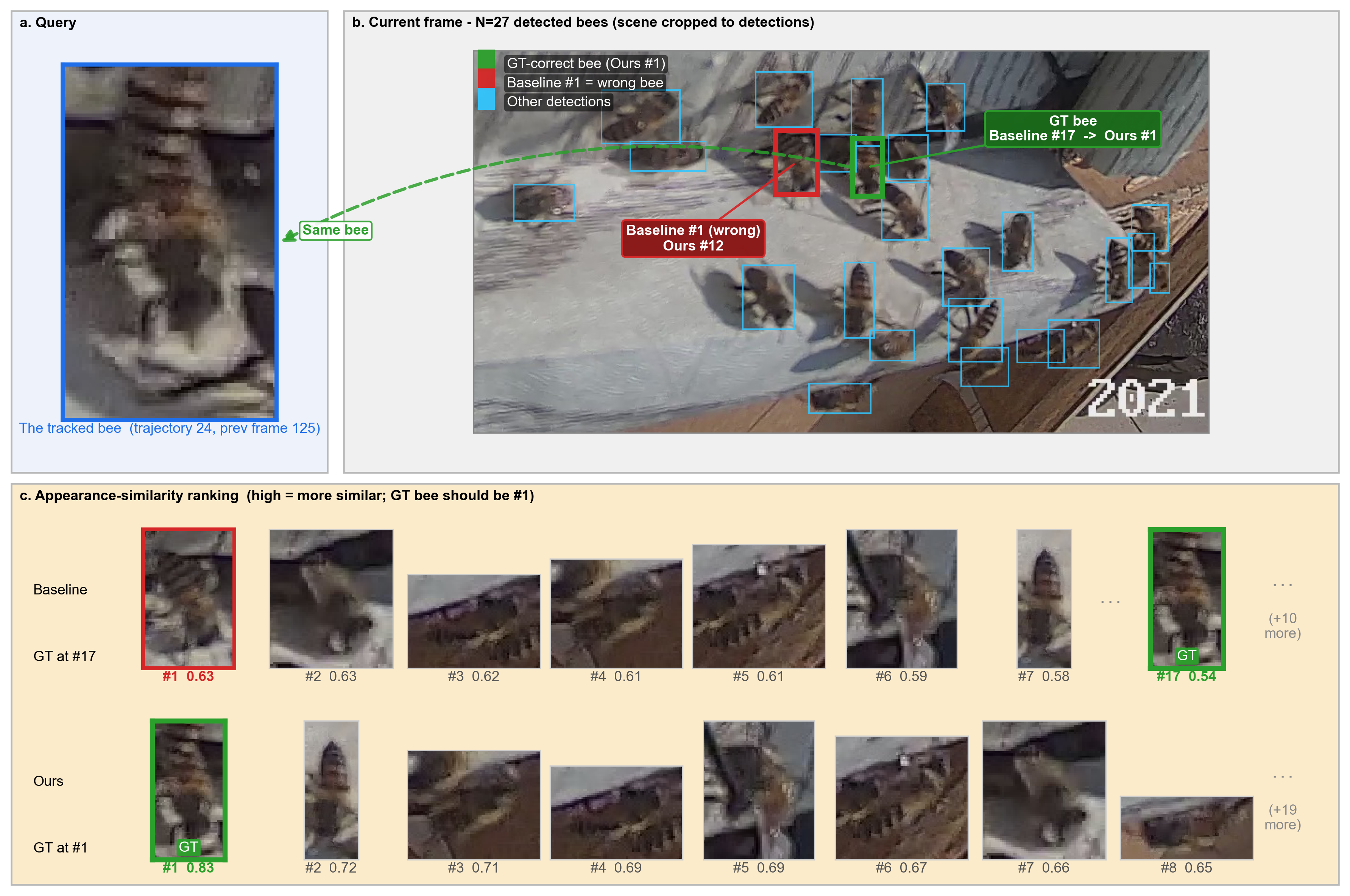}
\caption{The learned FCAE + CAS model moves the true candidate from rank 17 to rank 1 among 27 detections in BEE24-13.}
\label{fig:case-bee24-13}
\end{figure*}
\subsection{Ablation Study}
\label{sec:ablation}
\noindent\textbf{Component-wise ablation.} Table~\ref{tab:ablation} isolates the effects of candidate-set association training and commonality suppression. Baseline uses TrackTrack with the original cosine appearance distance. Baseline + Episode changes only the training objective, replacing instance-level re-ID training with trajectory-to-frame candidate-set association training. Baseline + Episode + Mean + CAS adds the parameter-free set-mean direction using Eq.~\eqref{eq:b_mean} as a commonality-suppression control. The final variant jointly learns FCAE and CAS.

Relative to Baseline, episode training increased MOTA from 65.7 to 67.5 and AssR from 54.9 to 61.9. It also reduced FN, IDs, and Frag. These changes show that candidate-set training improved association recall and trajectory continuity. However, IDF1 and AssA decreased to 63.1 and 44.4, while FP increased to 14856. Episode training therefore did not preserve association precision across all metrics when used alone.

The parameter-free Mean + CAS control increased HOTA, IDF1, and AssA to 48.6, 64.3, and 46.0 relative to the episode-only variant and reduced FP to 8409. This pattern is consistent with more selective association. Its AssR decreased to 57.5 and IDs increased to 542, indicating a trade-off between precision and association recall.

The learned FCAE + CAS variant achieved the highest HOTA, IDF1, and AssA and the lowest IDs among the four configurations. Relative to Baseline, HOTA increased from 48.1 to 49.2, IDF1 from 64.1 to 64.9, AssA from 45.9 to 46.6, and AssR from 54.9 to 61.7, while IDs decreased from 606 to 437. Relative to the episode-only variant, learned commonality suppression retained most of the AssR gain while recovering IDF1 and AssA and further reducing IDs.

\medskip
\noindent\textbf{Candidate discrimination analysis.}
We next measured the per-query similarity gap, defined as the score of the true candidate minus that of the highest-scoring incorrect candidate in the same frame. A larger positive gap implies stronger discrimination between the correct match and its hardest competing candidate. The mean gap increased from 0.1105 for Baseline to 0.2261 for the final method, while the distribution shifted consistently toward larger values (Fig.~\ref{fig:similarity-gap}). This indicates that the learned appearance representation produces more reliable within-frame candidate separation.

The candidate-ranking diagnostics in Fig.~\ref{fig:candidate-ranking} support the same conclusion. Specifically, each active trajectory is treated as a query, and all candidate detections in the current frame are ranked according to their association scores. The reported rank corresponds to the position of the ground-truth matching detection among all candidates. Rank-1 increased from 86.6\% to 88.2\%, Top-3 from 94.1\% to 95.1\%, and Top-5 from 96.4\% to 97.1\%. The mean rank of the true candidate decreased from 1.510 to 1.425, a reduction of 5.6\%. Together, the gap and ranking analyses show that the method improves within-frame candidate ordering.

Overall, video-level episode training narrowed the training--inference mismatch by optimizing each trajectory template against its same-frame candidate set. FCAE and CAS further improved this ranking by suppressing an estimated shared appearance component.

\subsection{Sensitivity Analysis}
\label{sec:sensitivity}
We varied the injection weight \(c\) from 0 to 0.50 while keeping the total appearance weight at 0.50, as defined in Eq.~\eqref{eq:blended-cost}. Within the episode-trained model, \(c=0\) used only the raw cosine distance, whereas \(c=0.50\) used only the set-relative distance. The dashed lines in Fig.~\ref{fig:weight-sensitivity} denote the original TrackTrack baseline.

At \(c=0\), episode training already improved several metrics over the original baseline. As \(c\) increased, HOTA generally increased and IDs decreased, while MOTA and AssR remained stable. Across the sweep, HOTA, MOTA, and AssR remained above their baseline values, and IDs remained below the baseline count. These trends indicate that the set-relative score provided consistent association gains over the tested range.

We selected \(c=0.50\) because it removes the mixing ratio at inference and achieved the lowest IDs while remaining close to the best HOTA value in the sweep. This full-replacement setting was used for the final model.

\subsection{Comparison with Existing Trackers}
\label{sec:sota}
Under the shared-detection protocol, VLA-ReID achieved the highest HOTA, MOTA, AssR, AssA, and IDF1 and the lowest IDs among the compared methods (Table~\ref{tab:sota}). It obtained 49.2 HOTA, 67.0 MOTA, 61.7 AssR, 46.6 AssA, and 64.9 IDF1, with 437 identity switches.

Relative to TrackTrack under the same detections and tracking pipeline, VLA-ReID improved HOTA, MOTA, and IDF1 by 1.1, 1.3, and 0.8 points. AssA and AssR increased by 0.7 and 6.8 points, respectively. IDs decreased from 606 to 437, a reduction of 28\%. These gains show that replacing only the appearance term improved association quality in the high-similarity BEE24 setting.

\subsection{Illustrative Case Analysis}
\label{sec:cases}
We used three selected frames to illustrate how commonality suppression changes within-frame candidate ranking. Each case contains a trajectory query, the current-frame detections, and the rankings produced by the baseline and the evaluated commonality-suppression variant. Green denotes the true candidate, red denotes the baseline's highest-ranked incorrect candidate, and blue denotes other detections. The first two figures visualize the parameter-free control, whereas the third visualizes the learned final model.

In BEE24-34, the baseline ranked a look-alike first and the true candidate fifth. The parameter-free commonality-suppression control moved the true candidate to rank 1 among 33 detections (Fig.~\ref{fig:case-bee24-34}). This example illustrates how set-relative scoring can change the ordering of visually similar candidates in a dense frame.

In BEE24-36, the baseline ranked an incorrect candidate first and the true candidate second. The parameter-free control moved the true candidate to first and the baseline's top-ranked candidate to third (Fig.~\ref{fig:case-bee24-36}). The case shows a local ranking correction when the true candidate and a hard negative receive similar cosine scores.

In BEE24-13, the baseline ranked the true candidate 17th among 27 detections. The learned FCAE + CAS model moved it to rank 1 (Fig.~\ref{fig:case-bee24-13}). This selected example shows that the final model can recover the correct ranking under low-resolution and unstable-lighting conditions.

Across the three illustrative cases, commonality-suppressed scoring moved the true candidate to rank 1. This qualitative pattern is consistent with the aggregate similarity-gap and Rank-\(k\) diagnostics, although the cases are selected examples rather than an unbiased evaluation set.

\section{Conclusion}
\label{sec:conclusion}

In this paper, we have presented VLA-ReID, a video-level association
framework for multi-object tracking in scenes with small, densely distributed,
and highly similar objects. Instead of learning appearance representations
through single-instance assignment, VLA-ReID takes the historical trajectories of
the current frame as queries and all its detections as candidates, directly
optimizing global association at the frame level and thereby aligning the
training objective with inference-time data association. On this basis, FCAE estimates the appearance direction common to all detections in the current frame, and CAS subtracts the component along this direction from both trajectory and detection features,
amplifying the discriminative differences among highly similar individuals
without additional annotations. Extensive experiments on the BEE24 dataset
demonstrate that VLA-ReID outperforms state-of-the-art trackers in terms of
HOTA, MOTA, AssR, AssA, and IDF1, while reducing identity switches by
28\%; ablation studies and sensitivity analyses further validate
the effectiveness of each module. In future work, we will extend the
video-level formulation to more trackers and to broader scenarios involving
highly similar objects.

\FloatBarrier
\bibliographystyle{IEEEtran}
\bibliography{ref}

\end{document}